\pdfoutput=1
\documentclass[11pt]{article}

\usepackage{EACL2023}

\usepackage{times}
\usepackage{latexsym}
\usepackage{graphicx}
\usepackage{amsmath, amssymb} % for mathematical notation
\usepackage{multirow}
\usepackage{lipsum} % allows to plot figure over 2 columns by using \begin{figure*}
\usepackage{float} % allows to use [H] in \begin{figure}

\usepackage[T1]{fontenc}
\usepackage[utf8]{inputenc}

\usepackage{microtype}

\usepackage{inconsolata}

\title{Identifying the Correlation Between Language Distance and Cross-Lingual Transfer in a Multilingual Representation Space}

\author{Fred Philippy\textsuperscript{1,2}\thanks{ \hspace{0.1cm} Research was conducted at Zortify.} \and Siwen Guo\textsuperscript{1} \and Shohreh Haddadan\textsuperscript{1} \\ \\ \textsuperscript{1}Zortify Labs, Zortify S.A. \\ 19, rue du Laboratoire L-1911 Luxembourg \\ \textsuperscript{2}SnT, University of Luxembourg \\ 29, Avenue J.F Kennedy L-1359 Luxembourg \\ \texttt{\{fred, siwen, shohreh\}@zortify.com}}

\begin{document}
\maketitle
\begin{abstract}
Prior research has investigated the impact of various linguistic features on cross-lingual transfer performance. In this study, we investigate the manner in which this effect can be mapped onto the representation space. While past studies have focused on the impact on cross-lingual alignment in multilingual language models during fine-tuning, this study examines the absolute evolution of the respective language representation spaces produced by MLLMs. We place a specific emphasis on the role of linguistic characteristics and investigate their inter-correlation with the impact on representation spaces and cross-lingual transfer performance.
Additionally, this paper provides preliminary evidence of how these findings can be leveraged to enhance transfer to linguistically distant languages.
\end{abstract}

\section{Introduction}
It has been shown that language models implicitly encode linguistic knowledge \citep{jawahar_what_2019, otmakhova_cross-linguistic_2022}. In the case of multilingual language models (MLLMs), previous research has also extensively investigated the influence of these linguistic features on cross-lingual transfer performance \citep{lauscher_zero_2020, dolicki_analysing_2021, de_vries_make_2022}. However, limited attention has been paid to the impact of these factors on the language representation spaces of MLLMs. 

Despite the fact that state-of-the-art MLLMs such as mBERT \citep{devlin_bert_2019} and XLM-R \citep{conneau_unsupervised_2020}, use a shared vocabulary and are intended to project text from any language into a language-agnostic embedding space, empirical evidence has demonstrated that these models encode language-specific information across all layers \citep{libovicky_language_2020, gonen_its_2020}. This leads to the possibility of identifying distinct monolingual representation spaces within the shared multilingual representation space \citep{chang_geometry_2022}.

Past research has focused on the cross-linguality of MLLMs during fine-tuning, specifically looking at the alignment of representation spaces of different language pairs \citep{singh_bert_2019, muller_first_2021}. Our focus, instead, is directed towards the absolute impact on the representation space of each language individually, rather than the relative impact on the representation space of a language compared to another one. Isolating the impact for each language enables a more in-depth study of the inner modifications that occur within MLLMs during fine-tuning. The main objective of our study is to examine the role of linguistic features in this context, as previous research has shown their impact on cross-lingual transfer performance. More specifically, we examine the relationship between the impact on the representation space of a target language after fine-tuning on a source language and five different language distance metrics. We have observed such relationships across all layers with a trend of stronger correlations in the deeper layers of the MLLM and significant differences between language distance metrics. 

Additionally, we observe an inter-correlation among language distance, impact on the representation space and transfer performance. Based on this observation, we propose a hypothesis that may assist in enhancing cross-lingual transfer to linguistically distant languages and provide preliminary evidence to suggest that further investigation of our hypothesis is merited.

\section{Related Work}
In monolingual settings, \citet{jawahar_what_2019} found that, after pre-training, BERT encodes different linguistic features in different layers. \citet{merchant_what_2020} showed that language models do not forget these linguistic structures during fine-tuning on a downstream task. Conversely, \citet{tanti_language-specificity_2021} have shown that during fine-tuning in multilingual settings, mBERT forgets some language-specific information, resulting in a more cross-lingual model.

At the representation space level, \citet{singh_bert_2019} and \citet{muller_first_2021} studied the impact of fine-tuning on mBERT's cross-linguality layer-wise. However, their research was limited to the evaluation of the impact on cross-lingual alignment comparing the representation space of one language to another, rather than assessing the evolution of a language's representation space in isolation.

%\citet{rajaee_how_2021} has investigated the impact of fine-tuning on the geometry, or more specifically the isotropy, of a language model's representation space.

\section{Methodology}
\subsection{Experimental Setup} \label{sec:experimental_setup}
In this paper, we focus on the effect of fine-tuning on the representation space of the 12-layer multilingual BERT model (\texttt{bert-base-multilingual-cased}). We restrict our focus on the Natural Language Inference (NLI) task and fine-tune on all 15 languages of the XNLI dataset \citep{conneau_xnli_2018} individually. We use the test set to evaluate the zero-shot cross-lingual transfer performance, measured as accuracy, and to generate embeddings that define the representation space of each language. More details on the training process and its reproducibility are provided in Appendix \ref{sec:reproducibility}.

\subsection{Measuring the Impact on the Representation Space}
We focus on measuring the impact on a language's representation space in a pre-trained MLLM during cross-lingual transfer. We accomplish this by measuring the similarity of hidden representations of samples from different target languages before and after fine-tuning in various source languages. For this purpose, we use the Centered Kernel Alignment (CKA) method \citep{kornblith_similarity_2019}\footnote{CKA is invariant to orthogonal transformations and thus allows to reliably compare isotropic but language-specific subspaces \citep{chang_geometry_2022}.}. When using a linear kernel, the CKA score of two representation matrices $X \in \mathbb{R}^{N \times m}$ and $Y \in \mathbb{R}^{N \times m}$, where $N$ is the number of data points and $m$ is the representation dimension, is given by
$$
CKA(X,Y) = 1 - \frac{\lVert XY^\intercal\rVert^2_F}{\lVert XX^\intercal\rVert_F \lVert YY^\intercal\rVert_F}
$$
where $\lVert \cdot \rVert_F$ is the Frobenius norm. \\

\paragraph{Notation}
We define $H_{S \rightarrow T}^i \in \mathbb{R}^{N \times m}$ as the hidden representation\footnote{We refer here to the hidden representation of the \texttt{[CLS]} token which is commonly used in BERT for classification tasks.} of $N$ samples from a target language $T$ at the $i$-th attention layer of a model fine-tuned in the source language $S$, where $m$ is the hidden layer output dimension. Similarly, we denote the hidden representation of $N$ samples from language $L$ at the $i$-th attention layer of a pre-trained base model (i.e. before fine-tuning) as  $H_L^i \in \mathbb{R}^{N \times m}$. More specifically, the representation space of each language will be represented by the stacked hidden states of its samples.

We define the impact on the representation space of a target language $T$ at the $i$-th attention layer when fine-tuning in a source language $S$ as follows:
$$
\Phi^{(i)}(S,T) = 1 - CKA\left(H_T^i,H_{S \rightarrow T}^i\right)
$$

\subsection{Measuring Language Distance}
In order to quantify the distance between languages we use three types of typological distances, namely the syntactic (SYN), geographic (GEO) and inventory (INV) distance, as well as the genetic (GEN) and phonological (PHON) distance between source and target language. These distances are pre-computed and are extracted from the URIEL Typological Database \citep{littell_uriel_2017} using \texttt{lang2vec}\footnote{\url{https://github.com/antonisa/lang2vec}}. For our study, such language distances based on aggregated linguistic features offer a more comprehensive representation of the relevant language distance characteristics. More information on these five metrics is provided in Appendix \ref{sec:lang_dist_appendix}.

\section{Correlation Analysis} \label{sec:correlation_analysis}
\begin{figure}[h!]
    \centering
    \includegraphics[width=0.45\textwidth]{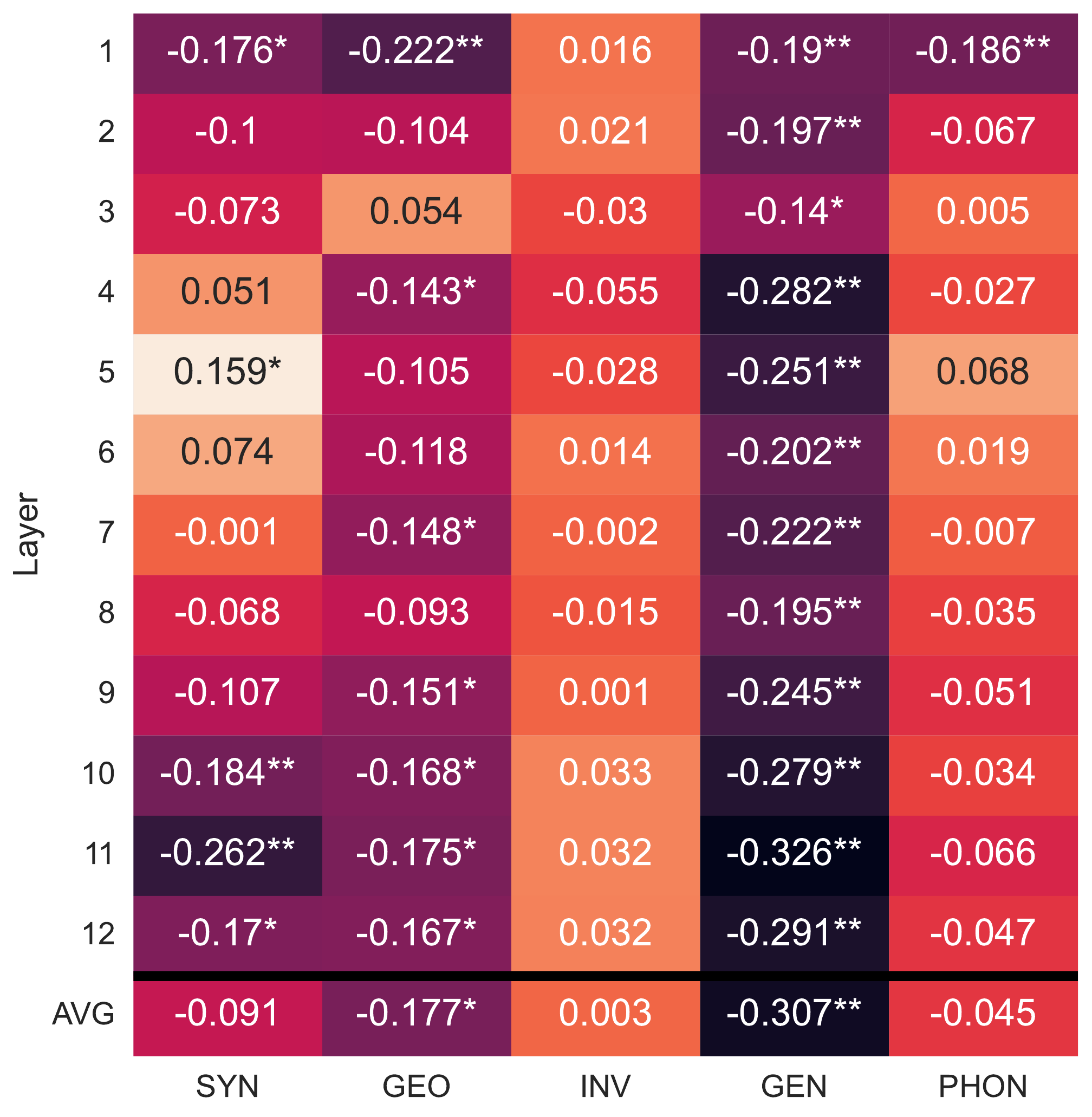}
    \caption{\textbf{Pearson correlation coefficient} between the \textbf{impact on a target language's representation space when fine-tuning in a source language} and different types of \textbf{linguistic distances between the source and target language} for each layer. Same source-target language pair data points were excluded in order to prevent an overestimation of effects. (${}^{*} p<0.05$, and ${}^{**} p<0.01$, two-tailed).}
    \label{fig:corelation_linguistic_impact}
\end{figure}

\paragraph{Relationship Between the Impact on the Representation Space and Language Distance.}
Given the layer-wise differences of mBERT's cross-linguality \citep{libovicky_language_2020, gonen_its_2020}, we measure the correlation between the impact on the representation space and the language distances across all layers. Figure \ref{fig:corelation_linguistic_impact} shows almost no significant correlation between representation space impact and \textbf{inventory} or \textbf{phonological} distance. \textbf{Geographic} and \textbf{syntactic} distance mostly show significant correlation values at the last layers. Only the \textbf{genetic} distance correlates significantly across all layers with the impact on the representation space.

\paragraph{Relationship Between Language Distance and Cross-Lingual Transfer Performance.}
Table \ref{tab:corr_langdist_perf} shows that all distance metrics correlate with cross-lingual transfer performance, which is consistent with the findings of \citet{lauscher_zero_2020}. Furthermore, we note that the correlation strengths align with the previously established relationship between language distance and representation space impact, with higher correlation values observed for syntactic, genetic, and geographic distance than for inventory and phonological distance. The exact zero-shot transfer results are provided in Figure \ref{fig:ZS_performance} in Appendix \ref{appendix:additional_figures}.

\begin{table}[h!]
    \centering
    %\small
    \begin{tabular}{lll}
     & Pearson & Spearman \\ \hline 
SYN  & $-0.3193^{**}$    & $-0.4683^{**}$     \\
GEO  & $-0.3178^{**}$    & $-0.3198^{**}$     \\
INV  & $-0.1706^{*}$     & $-0.1329^{*}$      \\
GEN  & $-0.3364^{**}$    & $-0.3935^{**}$     \\
PHON & $-0.2075^{**}$    & $-0.2659^{**}$     \\ \hline
\end{tabular}
    \caption{Pearson and Spearman \textbf{correlation coefficients} quantifying the relationship between \textbf{zero-shot cross-lingual transfer performance} and different \textbf{language distance metrics}. (${}^{*} p<0.05$, and ${}^{**} p<0.01$, two-tailed).}
    \label{tab:corr_langdist_perf}
\end{table}

\paragraph{Relationship Between the Impact on the Representation Space and Cross-Lingual Transfer Performance.}
In general, cross-lingual transfer performance clearly correlates with impact on the representation space of the target language, but this correlation tends to be stronger in the deeper layers of the model (Table \ref{tab:corr_impact_perform}).

\begin{table}[h!]
    \centering
    %\small
    \begin{tabular}{cll}
\hline
Layer   & Pearson         & Spearman      \\
\hline
1       & $0.2779^{*}$    & $0.3233^{*}$  \\
2       & $0.2456^{*}$    & $0.2639^{*}$  \\
3       & $0.5277^{*}$    & $0.5926^{*}$  \\
4       & $0.3585^{*}$    & $0.3411^{*}$  \\
5       & $-0.009$        & $0.0669$      \\
6       & $0.1033$        & $0.1969$      \\
7       & $0.2945^{*}$    & $0.3500^{*}$  \\
8       & $0.3004^{*}$    & $0.3517^{*}$  \\
9       & $0.4209^{*}$    & $0.4583^{*}$  \\
10      & $0.6088^{*}$    & $0.6532^{*}$  \\
11      & $0.7110^{*}$    & $0.7525^{*}$  \\
12      & $0.5731^{*}$    & $0.5901^{*}$  \\
\hline \hline
All     & $0.4343^{*}$    & $0.5026^{*}$  \\
\hline
\end{tabular}
    \caption{\textbf{Pearson correlation} coefficients between \textbf{cross-lingual transfer performance} and the \textbf{impact on the representation space of the target language}. (${}^{*} p<0.01$, two-tailed).}
    \label{tab:corr_impact_perform}
\end{table}

\section{Does Selective Layer Freezing Allow to Improve Transfer to Linguistically Distant Languages?} \label{sec:layer_freezing}

\begin{table*}[]
    \small
    \centering
    \renewcommand{\arraystretch}{1.5}

\begin{tabular}{cc|ccccc|c}
\multirow{2}{*}{Exp.} & \multirow{2}{1cm}{\centering Frozen Layers} & \multirow{2}{*}{SYN} & \multirow{2}{*}{GEO} & \multirow{2}{*}{INV} & \multirow{2}{*}{GEN} & \multirow{2}{*}{PHON} & \multirow{2}{*}{CLTP} \\
& & & & & & & \\ \hline
\textbf{} & \textbf{} & \textbf{-0.7354} & \textbf{-0.5109} & \textbf{-0.4907} & \textbf{-0.6116} & \textbf{-0.5776} & \textbf{66.70} \\
A & \{2\} & -0.7310 & -0.5109 & \underline{-0.4791}    & -0.6009 & -0.5791 & 66.53 \\
B & \{5\} & -0.7438 & -0.5053 & -0.4897 & -0.6148 & \underline{-0.5896} & 66.77 \\
C & \{1,2,6\} & \underline{-0.7325} & -0.5000 & \underline{-0,4846} & -0.6065 & \underline{-0.5666} & 66.75
\end{tabular}

% \begin{tabular}{cclllll|l}
% %\hline \\
% \multirow{4}{*}{Exp.} & \multirow{4}{1cm}{Frozen\\layers}   & \multicolumn{5}{c|}{\multirow{2}{*}{Correlation between transfer performance and linguistic distance}} & \multirow{4}{*}{CLTP} \\
% & & \multicolumn{5}{c|}{} & \\ \cline{3-7}
% & & SYN & GEO & INV & GEN & PHON & \\ \hline
% \multicolumn{1}{l|}{} & \multicolumn{1}{l|}{/} & \textbf{-0,7354} & \textbf{-0,5109} & \textbf{-0,4907} & \textbf{-0,6116} & \textbf{-0,5776} & 66,70 \\
% \multicolumn{1}{l|}{\#1} & \multicolumn{1}{l|}{\{2\}} & -0,7310 & -0,5109 & \underline{-0,4791} ($\uparrow$) & -0,6009 & -0,5791 & 66,53 \\
% \multicolumn{1}{l|}{\#2} & \multicolumn{1}{l|}{\{5\}} & -0,7438 & -0,5053 & -0,4897 & -0,6148 & \underline{-0,5896} ($\downarrow$) & 66,77 \\
% \multicolumn{1}{l|}{\#3} & \multicolumn{1}{l|}{\{1,2,6\}} & \underline{-0,7195} ($\uparrow$) & -0,5073 & \underline{-0,4909} (-) & -0,6025 & \underline{-0,5697} ($\uparrow$) & 66,75

% \end{tabular}
    \caption{Pearson \textbf{correlation coefficients} quantifying the relationship between \textbf{cross-lingual transfer performance} and different \textbf{language distance metrics} after freezing different layers during fine-tuning. The first row contains baseline values for full-model fine-tuning. The last column provides the average cross-lingual transfer performance (CLTP), measured as accuracy, across all target languages. English has been the only source language.}
    \label{tab:pilot_experiments}
\end{table*}

In the previous section we observed an inter-correlation between cross-lingual transfer performance, the linguistic distance between the target and source language, and the impact on the representation space. Given this observation, we investigate the possibility to use this information to improve transfer to linguistically distant languages. More specifically, we hypothesize that it may be possible to regulate cross-lingual transfer performance by selectively interfering with the previously observed correlations at specific layers. A straightforward strategy would be to selectively freeze layers, during the fine-tuning process, where a significant negative correlation between the impact on their representation space and the distance between source and target languages has been observed. By freezing a layer, we manually set the correlation between the impact on the representation space and language distance to zero, which may simultaneously reduce the significance of the correlation between language distance and transfer performance.

\citet{wu_beto_2019} already showed that freezing early layers of mBERT during fine-tuning may lead to increased cross-lingual transfer performance. With the same goal in mind, \citet{xu_soft_2021} employ meta-learning to select layer-wise learning rates during fine-tuning. In what follows, we will, however, not focus on pure overall transfer performance. Our approach is to specifically target transfer performance improvements for target languages that are linguistically distant from the source language, rather than trying to achieve equal transfer performance increases for all target languages.

\subsection{Experimental Setup}
For our pilot experiments, we focus on English as the source language. Additionally, we choose to carry out our pilot experiments on layers 1, 2, 5, and 6, as the representation space impact at these layers exhibits low correlation values with transfer performance (Table \ref{tab:corr_impact_perform}) and high correlations with different language distances (Figure \ref{fig:corelation_linguistic_impact_english} in Appendix \ref{appendix:additional_figures}). This decision is made to mitigate the potential impact on the overall transfer performance, which could obscure the primary effect of interest, and to simultaneously target layers which might be responsible for the transfer gap to distant languages. We conduct 3 different experiments aiming to regulate correlations between specific language distances and transfer performance. In an attempt to diversify our experiments, we aim to decrease the transfer performance gap for both a single language distance metric (Experiment \textbf{A}) and multiple distance metrics (Exp. \textbf{C}). Furthermore, in another experiment we aim at deliberately increasing the transfer gap (Exp. \textbf{B}).

%The capability to selectively regulate these correlations equates to the ability of impacting the transfer performance gap between linguistically distant languages.

\subsection{Results}
Table \ref{tab:pilot_experiments} provides results of all 3 experiments.
\paragraph{Experiment A.}
The \underline{$2^\text{nd}$ layer} shows a strong \underline{negative correlation} (-0.66) between representation space impact and \underline{inventory} distance to English. Freezing the $2^\text{nd}$ layer during fine-tuning has led to a less significant correlation between inventory distance and transfer performance (+0.0116).

\paragraph{Experiment B.}
The \underline{$5^\text{th}$ layer} shows a strong \underline{positive correlation} (0.499) between representation space impact and \underline{phonological} distance to English. Freezing the $5^\text{th}$ layer during fine-tuning has led to a more significant correlation between phonological distance and transfer performance (-0.012).

\paragraph{Experiment C.}
The \underline{$1^\text{st}$ layer}, \underline{$2^\text{nd}$ layer} and \underline{$6^\text{th}$ layer} show a strong \underline{negative correlation} between the impact on the representation space and the \underline{syntactic} (-0.618), \underline{inventory} (-0.66) and \underline{phonological} (-0.543) distance to English, respectively. Freezing the $1^\text{st}$, $2^\text{nd}$ and $6^\text{th}$ layer during fine-tuning has led to a less significant correlation of transfer performance with syntactic (+0.0029) and phonological (+0.011) distance.

\section{Conclusion}
In previous research, the effect of fine-tuning on a language representation space was usually studied in relative terms, for instance by comparing the cross-lingual alignment between two monolingual representation spaces before and after fine-tuning. Our research, however, focused on the absolute impact on the language-specific representation spaces within the multilingual space and explored the relationship between this impact and language distance. 
Our findings suggest that there is an inter-correlation between language distance, impact on the representation space, and transfer performance which varies across layers. Based on this finding, we hypothesize that selectively freezing layers during fine-tuning, at which specific inter-correlations are observed, may help to reduce the transfer performance gap to distant languages. Although our hypothesis is only supported by three pilot experiments, we anticipate that it may stimulate further research to include an assessment of our hypothesis.

\section*{Limitations}
It is important to note that the evidence presented in this paper is not meant to be exhaustive, but rather to serve as a starting point for future research. Our findings are based on a set of 15 languages and a single downstream task and may not generalize to other languages or settings. Additionally, the proposed hypothesis has been tested through a limited number of experiments, and more extensive studies are required to determine its practicality and effectiveness.

Furthermore, in our study, we limited ourselves to using traditional correlation coefficients, which are limited in terms of the relationships they can capture, and it is possible that there are additional correlations that could further strengthen our results and conclusions.

\section*{Ethics Statement}
This study was designed to minimize its environmental impact by reducing the amount of required computational resources to run our experiments. We are aware of the high energy consumption and carbon footprint associated with large-scale machine learning experiments and took steps to minimize these impacts.

Additionally, in this study, our objective was to address the performance gap in languages that are underrepresented in comparison to high-resource languages, rather than solely striving for performance enhancement.

% Entries for the entire Anthology, followed by custom entries
\bibliography{Bibliographies/export-data}
% \bibliography{anthology,custom}
\bibliographystyle{acl_natbib}

\appendix

\section{Technical Details}
\label{sec:reproducibility}
\subsection{Data}
We perform our experiments on the XNLI \citep{conneau_xnli_2018} dataset\footnote{\url{https://github.com/facebookresearch/XNLI}}. The dataset contains 392.702 train, 2.490 validation and 5.010 test samples, derived from the English-only MultiNLI \citep{williams_broad-coverage_2018}, which have been translated to 14 more languages. The objective of the dataset is to evaluate a model's capability of classifying the relationship between two sentences, namely a premise and a hypothesis, as entailment, contradiction, or neutral. 

The dataset has been released under a \textit{Creative Commons Attribution Non Commercial 4.0 International}\footnote{\url{https://creativecommons.org/licenses/by-nc/4.0/}} license (CC BY-NC 4.0).

\subsection{Model}
We use the base cased multilingual BERT \citep{devlin_bert_2019} model, which has 12 attention heads and 12 transformer blocks with a hidden size of 768. The dropout probability is 0.1. The model has 110M parameters and covers 104 languages. Its vocabulary size is about 120k.

\subsection{Training}
We fine-tune the models using the HuggingFace Transformers \citep{wolf_transformers_2020} and PyTorch \citep{paszke_pytorch_2019} frameworks. We use AdamW \citep{loshchilov_decoupled_2019} as an optimizer, with $\beta_1=0.9$, $\beta_2=0.999$, $\epsilon=1e^{-8}$. We train for 3 epochs with a batch size of 32 and an initial learning rate of $2e^{-5}$ with linear decay. Full model fine-tuning on a single language took about 2.5 hours on a single $\text{NVIDIA}^\text{\textregistered}$ V100 GPU. Total GPU hours for all 18 fine-tuned models (15 and 3 in Sections \ref{sec:correlation_analysis} and \ref{sec:layer_freezing} respectively was about 45 hours.

In order to minimize computational costs and reduce our environmental impact, we chose not to conduct a full hyper-parameter search and instead used the fixed values reported in Section \ref{sec:experimental_setup}.

For reproducibility, our code is provided here: \\ \url{https://github.com/fredxlpy/CrossLingualSpaceImpactAnalysis}.

\section{Additional Information on Language Distance Metrics} \label{sec:lang_dist_appendix}
%For all 5 language distance metrics we use the pre-computed distances provided by lang2vec.
We used the following lang2vec distances:
\begin{enumerate}
    \item \textbf{Syntactic Distance} is the cosine distance between the syntax feature vectors of languages, sourced from the World Atlas of Language Structures.\footnote{\url{https://wals.info}} (WALS) \citep{dryer_wals_2013}, Syntactic Structures of World Languages\footnote{\url{http://sswl.railsplayground.net/}} (SSWL) \citep{collins_syntactic_2011} and Ethnologue\footnote{\url{https://www.ethnologue.com/}} \citep{lewis_ethnologue_2015}.
    \item \textbf{Geographic Distance} refers to the shortest distance between two languages on the surface of the earth's sphere, also known as the orthodromic distance.
    \item \textbf{Inventory Distance} is the cosine distance between the inventory feature vectors of languages, sourced from the PHOIBLE\footnote{\url{https://phoible.org/}} database \citep{moran_phoible_2019}.
    \item \textbf{Genetic Distance} is based on the Glottolog\footnote{\url{https://glottolog.org}} \citep{hammarstrom_glottolog_2015} tree of language families and is obtained by computing the distance between two languages in the tree.
    \item \textbf{Phonological Distance} is the cosine distance between the phonological feature vectors of languages, sourced from WALS and Ethnologue.
\end{enumerate}

%The values range from 0 to 1, where 0 indicates the minimum distance and 1 indicates the maximum distance.

\section{Additional Figures} \label{appendix:additional_figures}

\begin{figure*}[ht]
    \centering
    \includegraphics[width=0.4\textwidth]{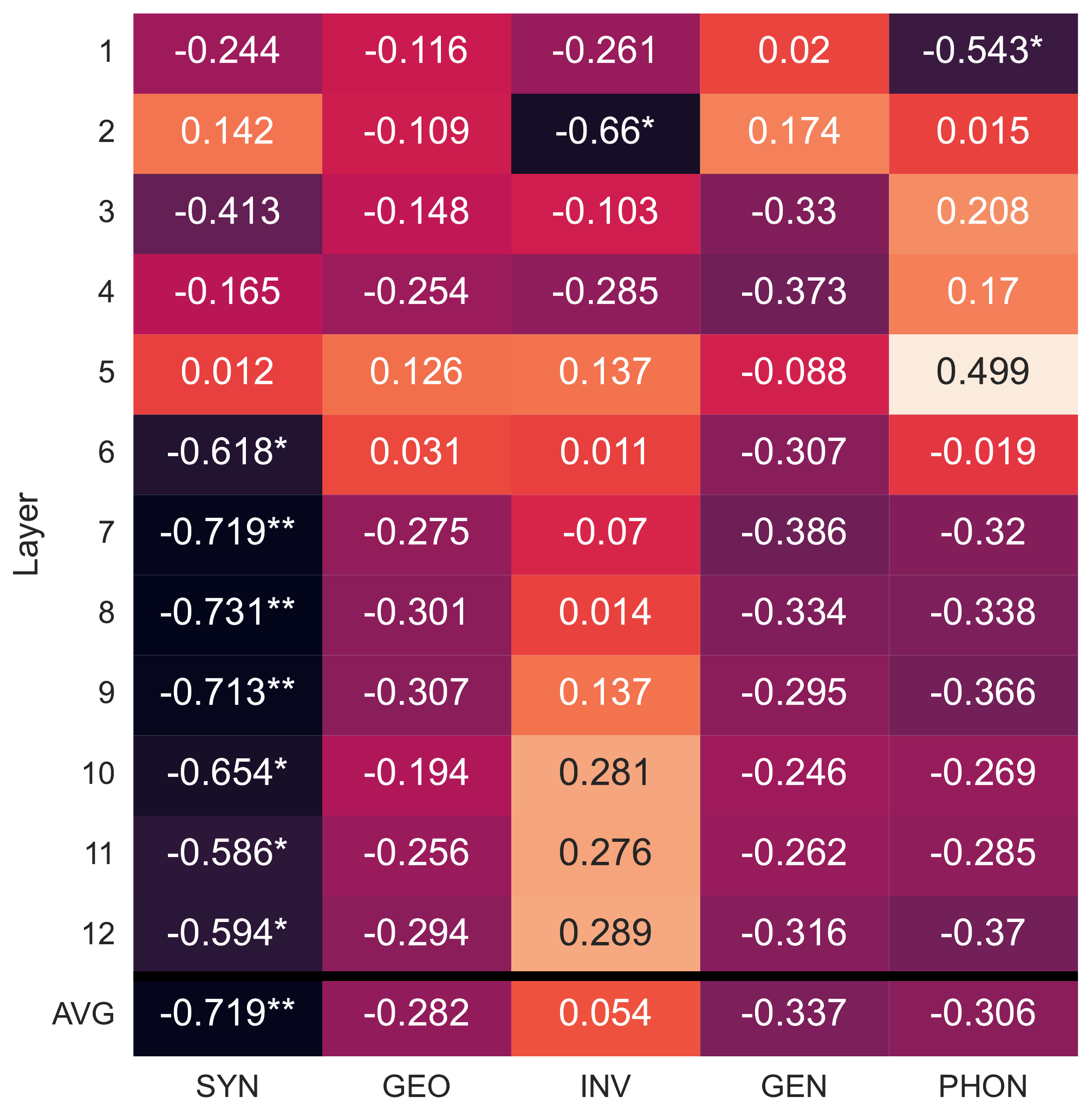}
    \caption{Pearson correlation coefficients between the impact on the representation space and different types of linguistic distances (with English as the only source language). (${}^{*} p<0.05$, and ${}^{**} p<0.01$, two-tailed).}
    \label{fig:corelation_linguistic_impact_english}
\end{figure*}

Figure \ref{fig:corelation_linguistic_impact_english} provides \textbf{Pearson correlation coefficients} between the \textbf{impact on the target language representation space} when fine-tuning in \textbf{\underline{English}} and different types of \textbf{linguistic distances between English and the target language} for each layer. English-English data points were excluded in order to prevent an overestimation of effects.

\begin{figure*}[ht]
    \centering
    \includegraphics[width=0.95\textwidth]{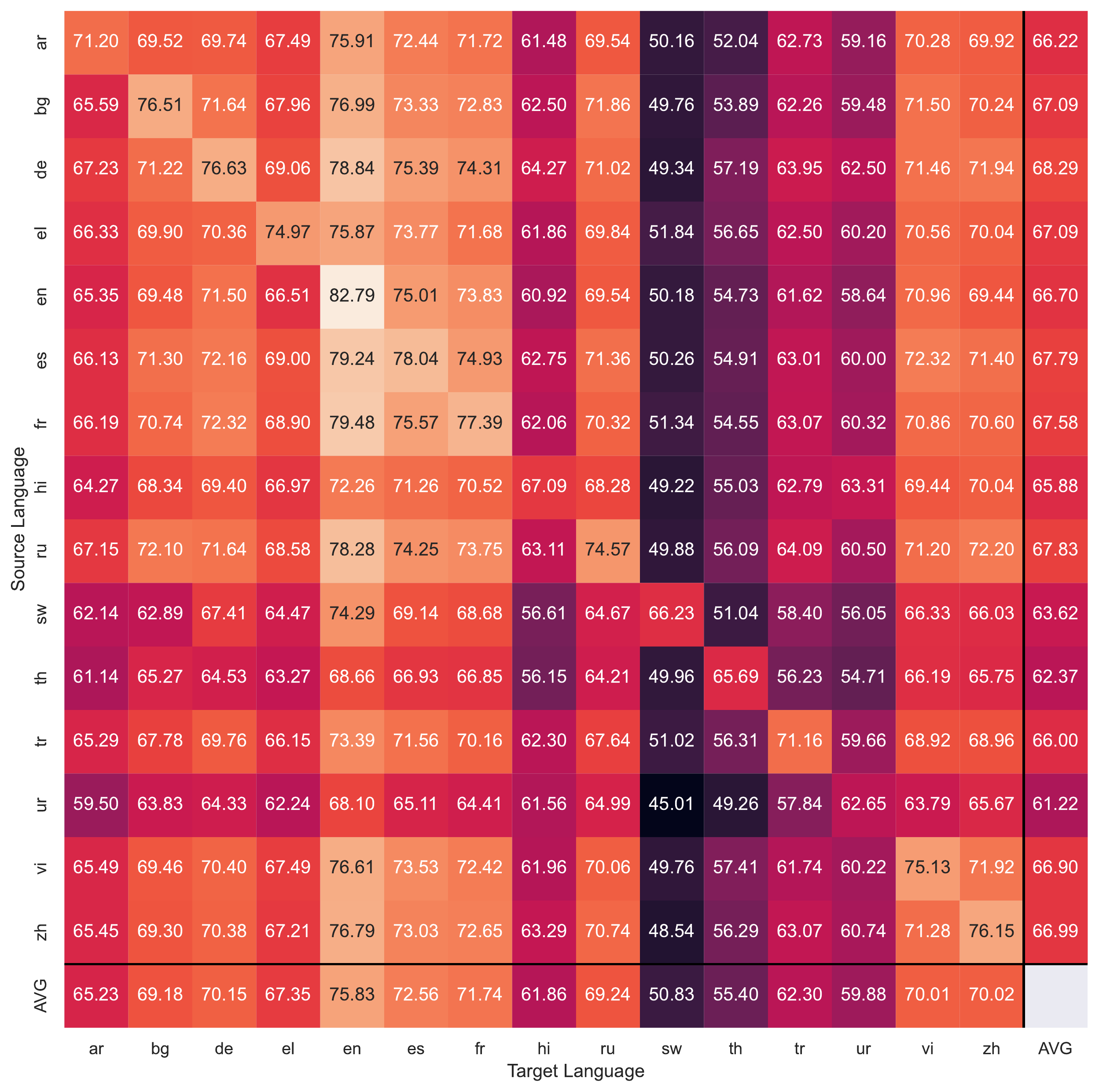}
    \caption{Cross-lingual zero-shot transfer results for XNLI}
    \label{fig:ZS_performance}
\end{figure*}

Figure \ref{fig:ZS_performance} contains the cross-lingual zero-shot transfer results. The numbers illustrated in the figure represent accuracies.

\end{document}